\documentclass[conference]{IEEEtran}
\IEEEoverridecommandlockouts
\usepackage{cite}
\usepackage{amsmath,amssymb,amsfonts}
\usepackage{algorithmic}
\usepackage{graphicx}
\usepackage{textcomp}
\usepackage{xcolor}
\usepackage{url}
\graphicspath{ {./} }
\def\BibTeX{{\rm B\kern-.05em{\sc i\kern-.025em b}\kern-.08em
    T\kern-.1667em\lower.7ex\hbox{E}\kern-.125emX}}
\begin{document}

\title{Deep Q-Network for Angry Birds\\
\thanks{This Research was funded by the Czech Science Foundation (project no. 17-17125Y).}
}

\author{\IEEEauthorblockN{1\textsuperscript{st} Ekaterina Nikonova}
\IEEEauthorblockA{\textit{Research School of Computer Science} \\
\textit{The Australian National University}\\
Canberra, ACT, Australia \\
ekaterina.nikonova@anu.edu.au
}
\and
\IEEEauthorblockN{2\textsuperscript{nd} Jakub Gemrot}
\IEEEauthorblockA{\textit{Faculty of Mathematics and Physics} \\
\textit{Charles University}\\
Prague, Czech Republic \\
gemrot@gamedev.cuni.cz}
}

\maketitle

\begin{abstract}
Angry Birds is a popular video game in which the player is provided with a sequence of birds to shoot from a slingshot. The task of the game is to destroy all green pigs with maximum possible score. Angry Birds appears to be a difficult task to solve for artificially intelligent agents due to the sequential decision-making, non-deterministic game environment, enormous state and action spaces and requirement to differentiate between multiple birds, their abilities and optimum tapping times. We describe the application of Deep Reinforcement learning by implementing Double Dueling Deep Q-network to play Angry Birds game. One of our main goals was to build an agent that is able to compete with previous participants and humans on the first 21 levels. In order to do so, we have collected a dataset of game frames that we used to train our agent on. We present different approaches and settings for DQN agent. We evaluate our agent using results of the previous participants of AIBirds competition, results of volunteer human players and present the results of AIBirds 2018 competition.
\end{abstract}


\section{Introduction}
Angry Birds has been one of the most popular video games for a period of several years. The main goal of the game is to kill all green pigs on the level together with applying as much damage as possible to the surrounding structures. The player is provided with a sequence of (sometimes) different birds to shoot from a slingshot. Usually, pigs are hidden inside of complex structures. The game requires a player to find and destroy some weak points of the structure such as its supports or hit a dynamite hidden inside them. Angry Birds was written using Box2D open source physics library and all objects in its game environment are following the laws of Newton Physics (in 2D). 
As the game is played by many, it seems that Angry Birds can be easily understood by almost everyone: from kids to adults. However, the task of playing the game is a challenge for the artificially intelligent agents, conversely to human players, for whom reasoning and planning in complex physical 2D world appears to be easy. This is due to a number of facts. Firstly, this game has a large number of possibilities of actions and nearly infinite amount of possible levels, which makes it difficult to use simple state space search algorithms for this task. Secondly, the game requires a planning of sequence of actions, which are related to each other or finding single precise shot. For example, a poorly chosen first action can make a level unsolvable by blocking a pig with a pile of objects. Therefore, to successfully solve the task, a game agent should be able to predict or simulate the outcome of it is own actions a few steps ahead.

\section{Related work}\label{related_work}
Back in 2012, the first Angry Birds AI competition (called AIBirds) was held \cite{firstcomp}. Many agents have been created since then. In this section we describe two agents which implementation details were publicly disclosed, and which performed well in the competition. One of the best agent that has been created during AIBirds competition was Datalab Birds’ 2014 made by a team from Czech Technical University in Prague. This agent still holds 3rd place in overall benchmark of all participating agents so far. As they describe in their paper \cite{datalab}, their main idea was to build a planning agent that decides which strategy to play for the current move based on the environment, possible trajectories and birds available. The agent always plans only one move ahead. The next move is chosen by taking a strategy with maximum estimated utility.

The second participant that we have looked at is Eagle’s Wings artificial player made by University of Waterloo and Zazzle Inc. for 2017 Angry Birds AI Competition. Their agent currently holds 16th place in overall benchmark. According to the description of their agent, it uses a simple multi-strategy affordance based on structural analysis with a manually tuned utility to decide between the strategies \cite{eaglewings}. This form of analysis gives multiple decisions and a manually tuned utility function predicting decisions' values. They were using the machine learning method called xgboost to learn the utility function of each decision.

In this paper, we are adapting Deep Q-networks as used to solve Atari games \cite{2013arXiv1312.5602M}. We hoped our agent to find weak structural points while maximizing outcome of each bird shot in Angry Birds game using reinforcement learning.

\section{Background}
In order to solve a sequential decision problem such as the Angry Birds game, we consider the game environment $\epsilon$ over discrete time steps. At each time step $t$ agent receives an observation $s_t$ and chooses an action from a set of available actions $a_t \in A = \{a_1...a_n\}$ for which it then receives reward $r_{t+1}$ from the environment. The agent goal then is to maximize the expected discounted reward following the policy $\pi$ \begin{equation}
Q^{\pi}(s,a) = \mathbb{E}[r_{t+1} + \gamma r_{t+2} + ...|s,a] \label{q_value}
\end{equation}

In \eqref{q_value}, $s$ is current state, $a$ is a chosen action, $r_{t+1}$ is the reward and $\gamma \in [0,1]$ is a discount, which specifies the importance of the rewards further into future.
We now define our optimal Q value to be
\begin{equation}
Q^*(s,a) = max_{\pi}Q^{\pi}(s,a)\label{q_value_opt}
\end{equation}
Where our optimal policy can be retrieved from \eqref{q_value_opt} by choosing action of the highest value in each state.

\section{Deep Q-Networks - Theory}
To approximate the optimal action-value function, we use deep neural network as the nonlinear function approximator. For this, we define an approximate value function 
\begin{equation} 
Q(s,a,\theta_i) \label{act_value}
\end{equation}

In \eqref{act_value}, $\theta_i$ are the weights of the Q-network at i-th iteration. 

As stated in the original paper \cite{hassabis:hlcontrol}, reinforcement learning algorithm that is used together with nonlinear function approximator such as neural networks can become unstable or even diverge due to the following problems: a) the correlation in the sequence of observations, b) from correlations between Q values and target values $ r_t + \gamma max_{a}Q(s_t,a) $ and c) by having a policy that is extremely sensitive to changes of Q value. 

Deep Q-learning is addressing the first problem by technique called experience replay. It is implemented by storing and later sampling the observations, which agent has experienced, randomly. This technique removes the correlation between the sequences of the observations by randomizing over the collected data. We define the experience to be 
\begin{equation} 
e_t=(s_t,a_t,r_{t+1},s_{t+1}) \label{exp}
\end{equation}
In \eqref{exp}, $s_t$ is a state at time $t$, $a_t$ is an action taken at time $t$, $r_{t+1}$ is a reward received at time ${t+1}$ and $s_{t+1}$ is a state an agent ended up after taking an action $a_t$ in. We store the experiences in the experience set 
\begin{equation} 
 M=\{e_1,…,e_t\} \label{exp_set}
\end{equation}
We later sample the experience set $M$ for a minibatches to update Q values inside of a Q-network. 

In order to address the second problem, following loss function is used:
\begin{multline}
    L_i(\theta_i) = \mathbb{E}_{(s,a,r,s^{'})\sim U(M)} [ (r+\gamma max_{a^{'}}Q(s^{'},a^{'},\theta_i^{-})\\ -Q(s,a,\theta_i ))^2 ]\label{loss}
\end{multline}

In \eqref{loss}, $i$ is the iteration, $\gamma$ is a discount factor, $\theta_i$ are weights of so-called online Q-network and $\theta_i^{-}$ are weights of a target network or so-called offline Q-network. The target network is called offline since its weights are only updated every $C$ steps with online network weights, while online network is updating every iteration $i$.

We then define the target that is used by DQN to be: 
\begin{equation} 
 Y_t = r_{t+1} + \gamma max_{a} Q(S_{t+1}, a,\theta_t^{-} ) \label{dqn_target}
\end{equation}
In \eqref{dqn_target}, $\theta_t^{-}$ are weights of target Q-network and $\gamma$ is a discount factor.

\subsection{Double Deep Q-Networks}
The original DQN uses the max operator for both selection and evaluation of the action. According to the \cite{2015arXiv150906461V} this makes it more likely to select overestimated values, which results in overoptimistic value estimates. For this reason the Double Q-learning algorithm was invented. The general idea behind it is that instead of using the parameters of only one network we will use the first network to select the action and the second network to evaluate that decision. In a more formal way, we will learn two value functions such that for each update one set of weights is used to determine the greedy policy and another one to determine the value of that policy. Thus we can rewrite the original Q-learning target to be: 
\begin{equation} 
 Y_t= r_{t+1}+Q(S_{t+1},argmax_{a} Q(S_{t+1},a,\theta_t ),\theta^{'}_t)  \label{ddqn_target}
\end{equation}
In \eqref{ddqn_target}, $\theta_t$ are online weights and $\theta^{'}_t$ are the second set of weights. Note that we are using online weights to select the action and second set of weights to evaluate it. Both of the weights can be updated by swapping their roles at random \cite{hasselt:doubleq}. 
Fortunately for us, the same idea of reducing the overestimations can be applied to  Deep Q-learning as we can replace the update function with the following:
\begin{equation} 
 Y_t= r_{t+1}+\gamma Q(S_{t+1},argmax_{a} Q(S_{t+1},a,\theta_t ),\theta^{-}_t) \label{ddqn_update}
\end{equation}
In \eqref{ddqn_update}, $\theta_t$ are weights of the online Deep Q-network and $\theta^{-}_t$ are weights of the target (offline) Deep Q-network. In comparison to previous Double Q-learning formula second set of weights $\theta^{'}_t$ is now replaced with weights of offline network $\theta^{-}_t$. As in the original Double Q-learning we are using the same idea of using online set of weights to determine the greedy policy and offline weights to evaluate current greedy policy. The update function itself stays the same as we have defined before, we are replacing offline weights with online weights each $\tau$ steps, where $\tau$ is a hyperparameter to be chosen.

\subsection{Dueling Deep Q-Network Architecture}\label{dual_arch}
In some situations, the values of the different actions are very similar, and it is unnecessary to estimate the value of each action. In case of Angry Birds, sometimes player would end up with an unsolvable level because pig was blocked with a pile of objects. It does not matter which action to take in this state, since all of them will have a similar value and we only care about the value of the state itself. As an improvement for these types of situations, the idea of Dueling architecture was introduced \cite{2015arXiv151106581W}. To implement the Dueling Q-learning architecture, we will have to use two streams of fully connected layers. Both of the layers should be constructed in a way that they have the ability to provide separate estimates of the advantage and value functions. After we have separated the output of convolutional layer to two streams we have to combine them back in order to obtain our Q function back.
As for the beginning, we first define the value function to be:
\begin{equation} 
V^{\pi} (s)=\mathbb{E}_{a~\pi(s)} [Q^{\pi}(s,a)] \label{value_f}
\end{equation}
And our advantage function to be:
\begin{equation} 
A^{\pi}(s,a)=Q^{\pi} (s,a)-V^{\pi}(s) \label{action_f}
\end{equation}
In other words, the value function tells us how good the particular state is, while advantage function tells us how important each action is by subtracting the value of the state from the value of choosing a particular action $a$ in state $s$. From \eqref{value_f} and \eqref{action_f} we can now define our new dueling module for Deep Q-network. We are going to make our first fully connected layer to have a single value output $V(s,\theta,\beta)$ and our second fully connected layer to have an output $A(s,a,\theta,\alpha)$ of dimension $|A|$. Here $\alpha$ and $\beta$ are weights of the advantage and the value fully connected layers and $\theta$ is weights of convolutional layers. In order to combine those two values together and to obtain Q values back, we define our last module of the network to be the following:
\begin{multline}
Q(s,a,\theta,\alpha,\beta)= V(s,\theta,\beta)+( A(s,a,\theta,\alpha) \\
-\frac{1}{|A|} \sum_{a^{'}} A(s,a',\theta,\alpha)) \label{merge_step}
\end{multline}
Thus, we get two streams of estimates of value and advantage functions which we will be using with our previously defined Deep Q-network and Double Q-learning for our playing agent in order to beat Angry Birds game in the following section.

\section{Deep Q-Networks for AIBirds}

In order to apply Deep Q-network to Angry Birds we first need to define a) the state, b) actions, c) a Q-network architecture and d) a reward function.
\begin{figure*}[h]
\centering
\includegraphics[scale=0.15]{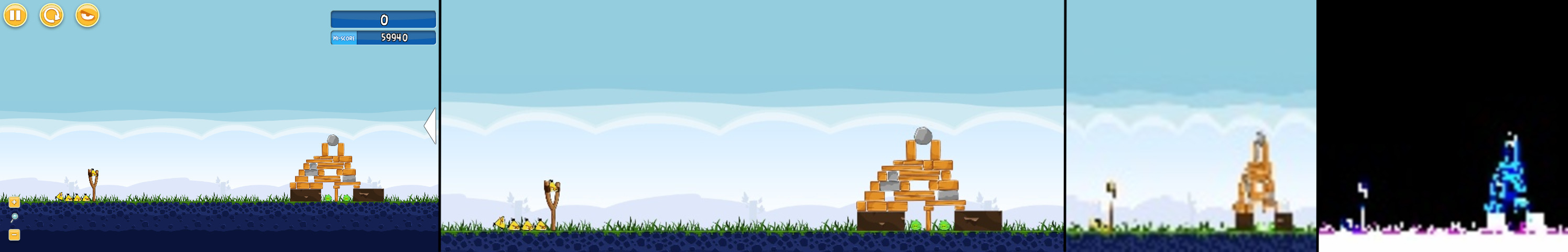}
\caption{Game frame at each state of preprocessing. From left to right: original captured frame, cropped, resized and normalized.}
\end{figure*}
The provided by AIBirds competition organizers software provides the ability to take screenshots of the game of size 840x480 pixels. On the Figure 1 we present the step by step process of preprocessing the image. We first crop captured frames by 770x310 pixels area that roughly captures the game play area and crops out the UI interface elements such as menu and restart level buttons. After the cropping we resize the cropped image to 84x84 pixels with RGB channels. At the end, we normalize the resulted image to have zero mean and unit norm and pass it as input to our Deep Q-network. 
\begin{figure}[h]
\centering
\includegraphics[scale=0.55]{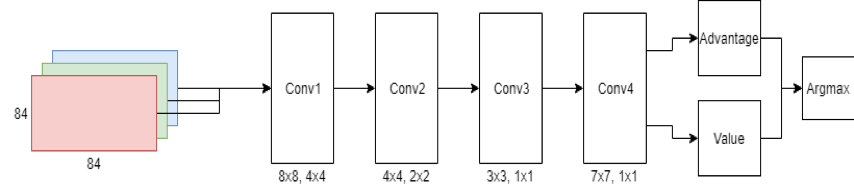}
\caption{Double Dueling Deep Q-Network architecture}
\end{figure}

We define the action to be $a\in \{0,1,2,3,…,90 \}$ where each discrete number represents the angle of the shot. We then find the final release point for a given angle using the provided software. To calculate the release point, trajectory module first finds the reference point for a sling and then calculates the release point using it. We decided to choose such a big action space since there are times where even 1 degree can make a huge difference. During our experiments the typical situation that we would be observing is that agent shots at 49 degrees, receives some small amount of points, but the next time it shots at 50 degrees and finishes the level with a new high score.

Our Deep Q-network architecture was based on the Google DeepMind Deep Q-network \cite{2013arXiv1312.5602M} that was used to play multiple Atari games. On the Figure 2 we present the architecture of our Deep Q-network. The model was consisted of 4 convolutional layers of kernel sizes 8x8, 4x4, 3x3, 7x7 and strides 4x4, 2x2, 2x2, 1x1 respectively. The last convolutional layer was followed by two flatten streams of value and advantages values that were later combined into final Q values as described in \ref{dual_arch}.

As for the reward function, we have experimented with two different versions. At first, our reward function was defined as follows:
\begin{equation} 
reward= 
\begin{cases}
       1,& \text{if } score > 3000 \\
     -1,& \text{otherwise} 
\end{cases} \label{reward_v1}
\end{equation}
 
Reward function \eqref{reward_v1} was using technique that is usually called reward clipping \cite{2016arXiv160207714V}. It squeezes the potentially unbounded score to two extremes of reward 1 for a “good” action and -1 for a “bad” action. With this reward function agent was able to perform quite well achieving on average the score of 850,000 (for the first 21 levels of “Poached Eggs” episode). However, with this reward function agent had no way to learn the important fluctuations between actions. For example, the action that obtained the score of 70,000 was not any more useful than action that obtained score of 3,001.
One of the main goals in Angry Birds besides winning the level, is to obtain a maximum possible score. Because of that, we had to change our reward function to somehow encapsulate the importance of the magnitude of the score without simply assigning the obtained score as our reward. 
Thus, we have redefined our reward function to be:
\begin{equation} 
reward=\frac{score}{maximum\_score\_of\_current\_level} \label{reward_v2}
\end{equation}

In \eqref{reward_v2}, $score$ is obtained score for some action $a$ at some state $s$, and $maximum~score~of~current~level$ is a maximum score in this level ever obtained either by one of our volunteers or by an agent itself. With this definition we were expecting agent to learn the importance of the score and eventually set new maximum scores for the levels by learning the most rewarding actions.

\subsection{Training set}

Our training set of levels consisted of 21 levels of “Poached Eggs” episode of Angry Birds Classic game. For the training process we have collected over 95,000 screenshots while agent was playing those 21 levels using greedy epsilon policy and approximately 20,000 using completely random policy. Thus, by mixing both data sets together, we had over 115,000 of the images that we used to train our network. The agent was waiting for 5 seconds before taking a screenshot of the game playing area. This small adjustment was needed in order to let physics to settle after the shockwave produce by the bird shot.

\subsection{Validation set}
Our validation set of levels consisted of 10 levels of "Poached Eggs" episode of Angry Birds Classic game. Since our agent was trained on levels with only red, blue, yellow birds we had to select the validation levels with no new birds. Our validation levels included levels 1,2,3,4,10 and 12 from the second page of the episode and levels 1,3,4 and 5 from third page of the episode. We have selected this set of levels since they were previously unseen by our agent and they did not contain any new birds. In particular, these levels did not contain the white bird. This bird was specifically eliminated from our level sets as it required agent to usually shoot above the structures and pick a particular tapping time to drop the bomb, which was different from all other birds. Levels from second and third page of the “Poached Eggs” episode are considered to be of a higher difficulty than those from the first page on which our agent was trained. Third page episodes are considered to be hard to complete even for humans. These levels usually require player to find some non-obvious weak point of the structure and to carefully plan actions ahead of a time. 

\section{Results}

\begin{figure}[h]
\centering
\includegraphics[scale=0.58]{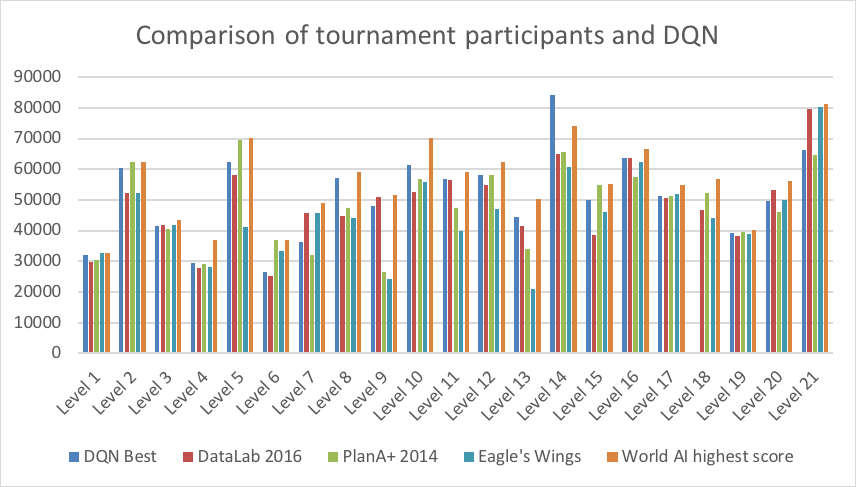}
\caption{Comparison of DQN agent to previous participants of AIBirds}
\end{figure}
On the Figure 3, we compare results of several participants of the competition on 21 levels. Our agent had only 1 try to complete each level and in case of a failure agent received 0 score for that level. For the comparison, we have selected results of Datalab 2016 agent which holds the 1st place in the tournament benchmark overall years and PlanA+ 2014 agent which holds the 2nd place. We have selected these two agents since they showed the highest scores on first 21 levels. For the better comparison, we also present overall highest achieved score for each level by artificial agents. 
Our agent was able to compete with some of the best agents ever presented on the AIBirds competition. Surprisingly enough, the agent was able to beat the high score for level 14. At first, we thought it was a mistake or a lucky outcome, however multiple reevaluation of the agent showed the same results as presented here. 

Comparing the results of different agents can help to understand which AI technique for solving Angry Birds game works better. However, what still reminds unclear after such a comparison is how well the agent compares to the human. For this paper, we have also conducted the Artificial Intelligence versus Human competition. The goal of the experiment was to achieve the maximum possible score for each level. Our participants were given an unlimited amount of time and tries. They were constantly replaying the levels, slowly achieving what they thought was their personal highest score. Once the participant decided that he or she achieved that maximum the experiment was stopped for him or her. We have also selected participants with different level of experiences. Thus, Human 1 and Human 3 claimed to have at least 4 years of experience in playing Angry Birds game. Human 3 has also claimed to play Angry Birds almost every day at the time of the experiment. Human 2 claimed to played Angry Birds a couple times a long time ago and thus did not have much experience. Human 4 claimed that she has played Angry Birds a lot in the past, but did not played much recently, thus we considered her to be close to the average player.

\begin{figure}[h]
\centering
\includegraphics[scale=0.58]{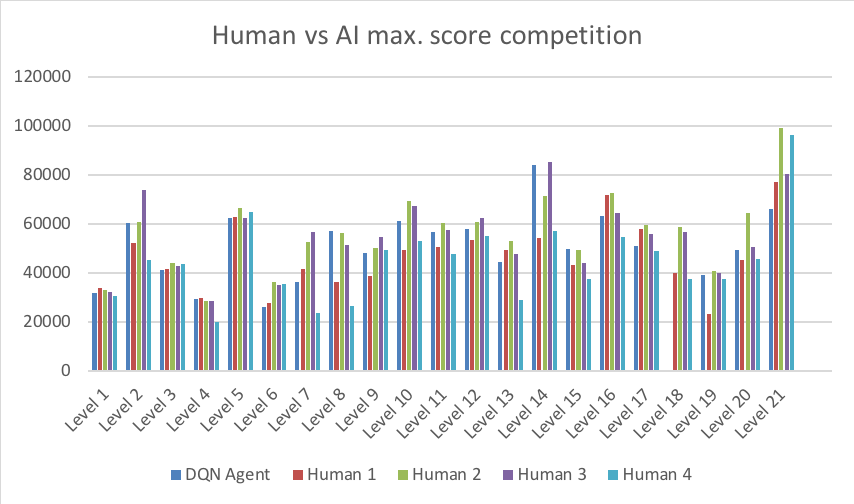}
\caption{Comparison of best DQN agent to human players}
\end{figure}
As we can see on the Figure 4, our agent was able to surpass one human but lost to others in a total sum of obtained scores for 21 levels. One of the main reasons agent has failed to obtain higher total score was due to its inability to beat level 18 in this particular setting. On the rest of the levels, agent was able to outperform some of the human players in some of the levels. Overall, our DQN agent was able to achieve 1,007,189 points on first 21 levels of the Poached Eggs episode. Comparing it to total scores of humans: 1,019,190, 983,840, 1191660 and 1,152,830 - it lost to humanity by losing in three out of fourth comparisons. Even in the level 14 where agent was able to beat the highest score ever achieved by artificially intelligent agents it lost to the third human player. Thus we can conclude that there are still place for the improvements for our agent. 

\begin{figure}[h]
\centering
\includegraphics[scale=0.58]{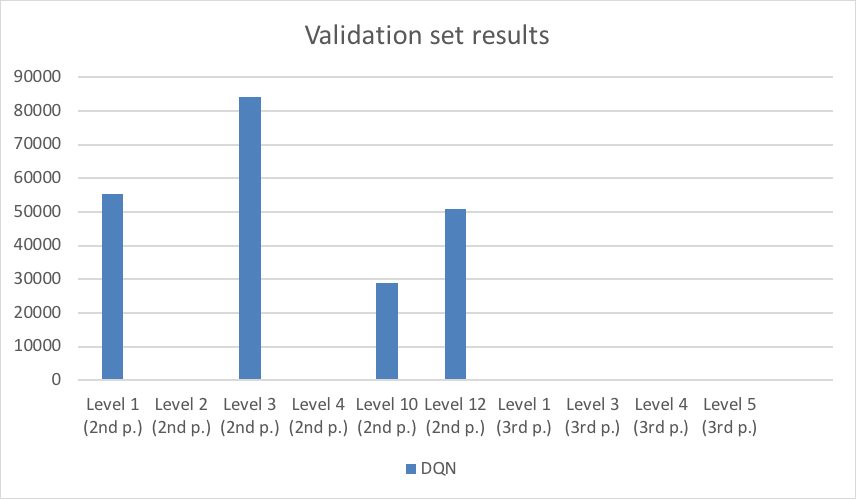}
\caption{DQN agent results on validation set. Here 2nd p. and 3rd p. means second and third pages of the "Poached Eggs" episode respectively.}
\end{figure}
On the Figure 5 we present the results of the agent on validation data set. As we were expecting, the agent was not able to complete most of the previously unseen levels. It was still surprising however, to see it completing some of the levels. From these results, we can conclude that our training data set did not include the whole variety of possible levels in the game, which lead to poor generalization. As a possible improvement, we could extend our training set to more difficult levels.

\subsection{Angry Birds AI Competition Results} \label{comp_results}
Beside the comparison of our agent to humans and previous participants, it was also presented on IJCAI-2018 Symposium on AI in Angry Birds and had participated in Angry Birds AI Competition \cite{aibirdscomp}. The competition itself was consisting out of three rounds: Quarter Final, Semi Final and Grand Final. In each round playing agents had to solve 8 previously unseen levels. As competition organizers have stated themselves, these levels were made with the intention to eliminate the agents that were using a simple brute-force algorithms and favor those who were using some artificial logic in order to pick the object to shoot at.
\begin{table}[htbp]
\caption{Angry Birds AI Competition Quarter Final Ranking}
\begin{center}
\begin{tabular}{|c|c|c|}
 \hline
 Rank & Team Name & Score \\
 \hline
\hline
1 & Eagle'sWing 2017 &	405,260	\\
\hline
2 & IHSEV &	398,480	 \\
\hline
3 & Eagle's Wing 2018 &	323,050	 \\
\hline
4 & BamBirds	& 290,020 \\
\hline
5 & PlanA+ &	272,060	 \\	 	 
\hline
6 & DQ-Birds	& 185,869 \\	 	 	 
\hline
7 & MetaBirds &	163,920	\\ 	 	 
\hline
8 & AngryHex	& 145,240 \\	 	 
\hline
9 & MYTBirds	& 49,900 \\
\hline
\end{tabular}
\label{results_table}
\end{center}
\end{table}

Table \eqref{results_table} shows us the total scores obtained by agents after all Quarter Finals have been completed \cite{aibirdscomp}. We decided to choose Quarter Final results as all of the presented agents were participating and all of them had to solve the same 8 levels over fixed time period of 30 minutes. Our agent (DQ-Birds) had shown the best result among all ever participated neural network-based agents as of 2018 by solving 3 out 8 levels and surpassing 2017 Semi-Finalist AngryHex. Besides our agent, there was another team (MYTBirds) which was also using neural networks in their agent. However, their agent was only able to solve 1 level out of 8. The rest of the agents were using different approaches which did not require to use neural networks. The winner of a Quarter Finals and later Grand Finals was a Eagle's Wing 2017 agent which was using  multi-strategy affordance, which we have mentioned in section \ref{related_work}.

\section{Discussion}
Angry Birds game is a challenging task for artificially intelligent playing agents that are using neural nets due to the fact that each level has some unique features which agent might never have seen before. As we have seen in \ref{comp_results}, despite the fact that our agent was able to master 21 levels from a training set and was able to solve previously unseen levels of a greater difficulty from a validation set, it still had a problem to solve all 8 levels during the competition. We could assume that exposing our agent to more levels could have lead to a better results, but we were unable to test this theory due to the inability to obtain an official level generator. On the other hand, another participant MYTBirds mentioned during their presentation on IJCAI-2018 Symposium on Angry Birds AI that they were using over 100 different levels to train their agent. However, they had to use a clone of the Angry Birds game due to the earlier mentioned problem of not having an official level generator. The clone of the game had a similar but slightly different physics engine compare to the original Angry Birds game. We assume, that using clone of the game might have affected the results of MYTBirds agent which would explain the low results shown during the competition.

Beside the problem of having a large number of possible levels, Angry Birds game has a continuous shooting angle space which is problematic for a neural network approximation using DQN method. This is due to the fact that in our approach we had a single output for each valid action, thus, for a continuous space we would had to define an arbitrary large output layer. To overcome this obstacle, we have defined the action space to be discrete. We defined the action to be $a\in \{0,1,2,3,…,90 \}$ with each discrete number representing the angle of the shot. By limiting the shooting angle in such a way, we were trying to reduce the chance of missing the optimal shot while keeping the size of output layer as low as possible. As a part of the experiment, we have also tried to reduce the action space to every 2 degrees, but it has dramatically reduced the overall agent's performance. One of the limits introduced with this action space, was agent inability to shot at points located lower than 0 degrees relatively to the center of the slingshot. However, this limitation did not significantly affect agent's performance on the first 21 levels since all of them did not require such shots.

On regards of hyper parameters of our network we have achieved the best results with the following setting: batch size = 32, learning rate = 0.00001, update rate = 4 and normalized reward function as was described in previous section.
\begin{figure}[h]
\centering
\includegraphics[scale=0.58]{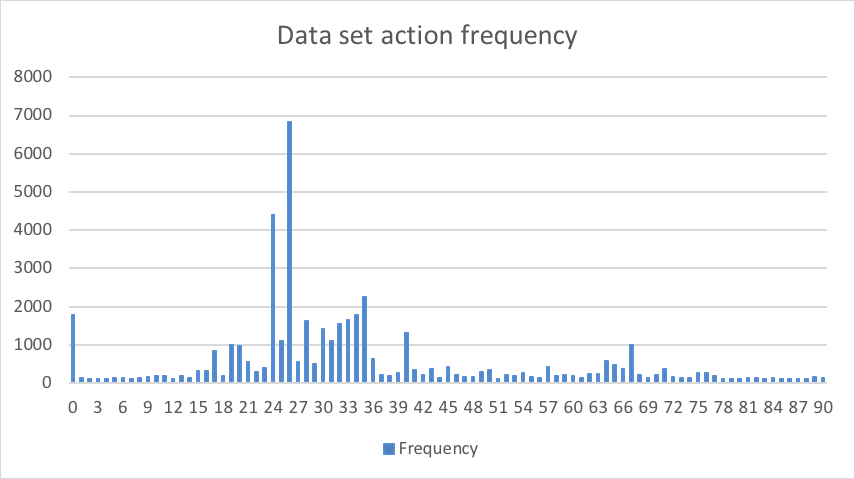}
\caption{Data distribution}
\end{figure}

On the Figure 6, one can see the distribution of actions overall collected data. Here we note that most of the actions in our data were distributed between 17 and 36 degrees, with strong outliers of 0, 40 and 67 degrees. Based on this data we were expecting the agent to learn the actions that closely resemble the graph above. For example, we were expecting that it would mostly shoot at the range of 24 to 40 degrees, as well as have some odd shots of lower and higher degrees in some levels. In fact, this was the case, as we can see in the action distribution over time on Figure 7.
\begin{figure}[h]
\centering
\includegraphics[scale=0.58]{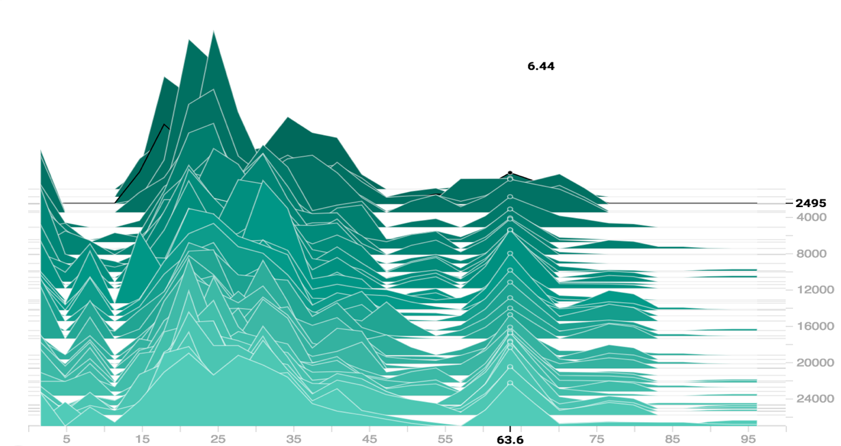}
\caption{Distribution of predicted actions over time. Vertical axis shows us the time, while horizontal axis shows the action (degree).}
\end{figure}

\section{Conclusion}
The Angry Birds game still remains a difficult task for artificially intelligent agents. We have presented a playing agent that was using Double Dueling Deep Q-Network in order to solve sequential decision problem in Angry Birds game. In order to train our agent, we have collected over 115,000 of game frames using greedy epsilon and partially random policy. One of the goals that we did not quite achieve in this work, is to outperform humans in Angry Birds game. Despite the fact we have come close to a human-level performance on selected 21 levels, we still lost to 3 out of 4 humans in obtaining a maximum possible total score. On a good side, our agent was able to learn to complete the levels using only one try. Another interesting point, is that most of the times it was using only one precise shot to some weak point which lead to the completion of the level.

As of the competition results, we were quite surprised that despite the fact that we have only used a small number of levels to train our agent, it was still able to solve 3 out of 8 levels and showed the highest result among all ever presented agents that used neural networks as part of their architecture.

Overall, while our agent had outperformed some of the previous and current participants of AIBirds competition and set new high-score for one of levels, there are still a lot of room for improvements. As an example, we could have experiment more with hyper parameters and definition of reward function. As a more radical improvement, we could have used any of the nowadays published enhancements for Deep Reinforcement Learning \cite{2017arXiv171002298H}. As our next step, we would like to train the agent on the much bigger number of Angry Birds levels.

\section*{Acknowledgements}

This Research was funded by the Czech Science Foundation (project no. 17-17125Y).

\end{document}